\documentclass[10pt,twocolumn,letterpaper]{article}

\usepackage{cvpr}
\usepackage[numbers,sort]{natbib}
\usepackage{times}
\usepackage{epsfig}
\usepackage{graphicx}
\usepackage{amsmath}
\usepackage{amssymb}
\usepackage{enumitem}
\usepackage{subcaption}
\usepackage{placeins}
\usepackage{siunitx}
\usepackage{pdfpages}

\usepackage{booktabs}
\usepackage{makecell}
\usepackage[table,dvipsnames]{xcolor}
\newcommand{\flow}[1]{\textcolor[rgb]{0.0,0.58,0.}{#1}}
\newcommand{\target}[1]{\textcolor[rgb]{1,0,0}{#1}}
\newcommand{\source}[1]{\textcolor[rgb]{0,0,1}{#1}}
\newcommand{\rgb}[1]{\textcolor[rgb]{0.5,0.5,1}{#1}}

\newcommand{\myparagraph}[1]{\vspace{1pt}\noindent \textbf{#1.}}

\usepackage[pagebackref=true,breaklinks=true,letterpaper=true,colorlinks,bookmarks=false]{hyperref}

\cvprfinalcopy % *** Uncomment this line for the final submission

\def\cvprPaperID{6128} % *** Enter the CVPR Paper ID here

\ifcvprfinal\fi
\begin{document}

%%%%%%%%% TITLE
\title{Multi-Modal Domain Adaptation for Fine-Grained Action Recognition}

\author{Jonathan Munro\\
University of Bristol\\
{\tt\small jonathan.munro@bristol.ac.uk}
\and
Dima Damen\\
University of Bristol\\
{\tt\small dima.damen@bristol.ac.uk}
}

\maketitle
\thispagestyle{empty}

%%%%%%%%% ABSTRACT
\begin{abstract}
Fine-grained action recognition datasets exhibit environmental bias, where multiple video sequences are captured from a limited number of environments.
Training a model in one environment and deploying in another results in a drop in performance due to an unavoidable domain shift.
Unsupervised Domain Adaptation (UDA) approaches have frequently utilised adversarial training between the source and target domains.
However, these approaches have not explored the multi-modal nature of video within each domain.
%In this work, we propose UDA through jointly optimising adversarial training with multi-modal self-supervised alignment.}
In this work we exploit the correspondence of modalities as a self-supervised alignment approach for UDA in addition to adversarial alignment (Fig.~\ref{fig:motivation}).

%We first show that modalities differ in the level of robustness to changes in the environment, tested on multiple kitchens in the 
We test our approach on three kitchens from our large-scale dataset, EPIC-Kitchens~\cite{damen2018scaling}, using two modalities commonly employed for action recognition: RGB and Optical Flow. We show that multi-modal self-supervision alone improves the performance over source-only training by 2.4\% on average. We then combine adversarial training with multi-modal self-supervision, showing that our approach outperforms other UDA methods by $3\%$.
% We employ late fusion of the two modalities commonly used in action recognition (RGB and Flow), with multiple domain discriminators, so alignment of modalities is jointly optimised with recognition. We test our approach on EPIC Kitchens~\cite{damen2018scaling}, proposing the first benchmark for domain adaptation of fine-grained actions. Our multi-modal method outperforms single-modality alignment as well as other alignment methods by up to~$3\%$.
\end{abstract}

%%%%%%%%% BODY TEXT
\section{Introduction}
\label{sec:intro}
\begin{figure}[t]
\centering
\includegraphics[width=1\linewidth]{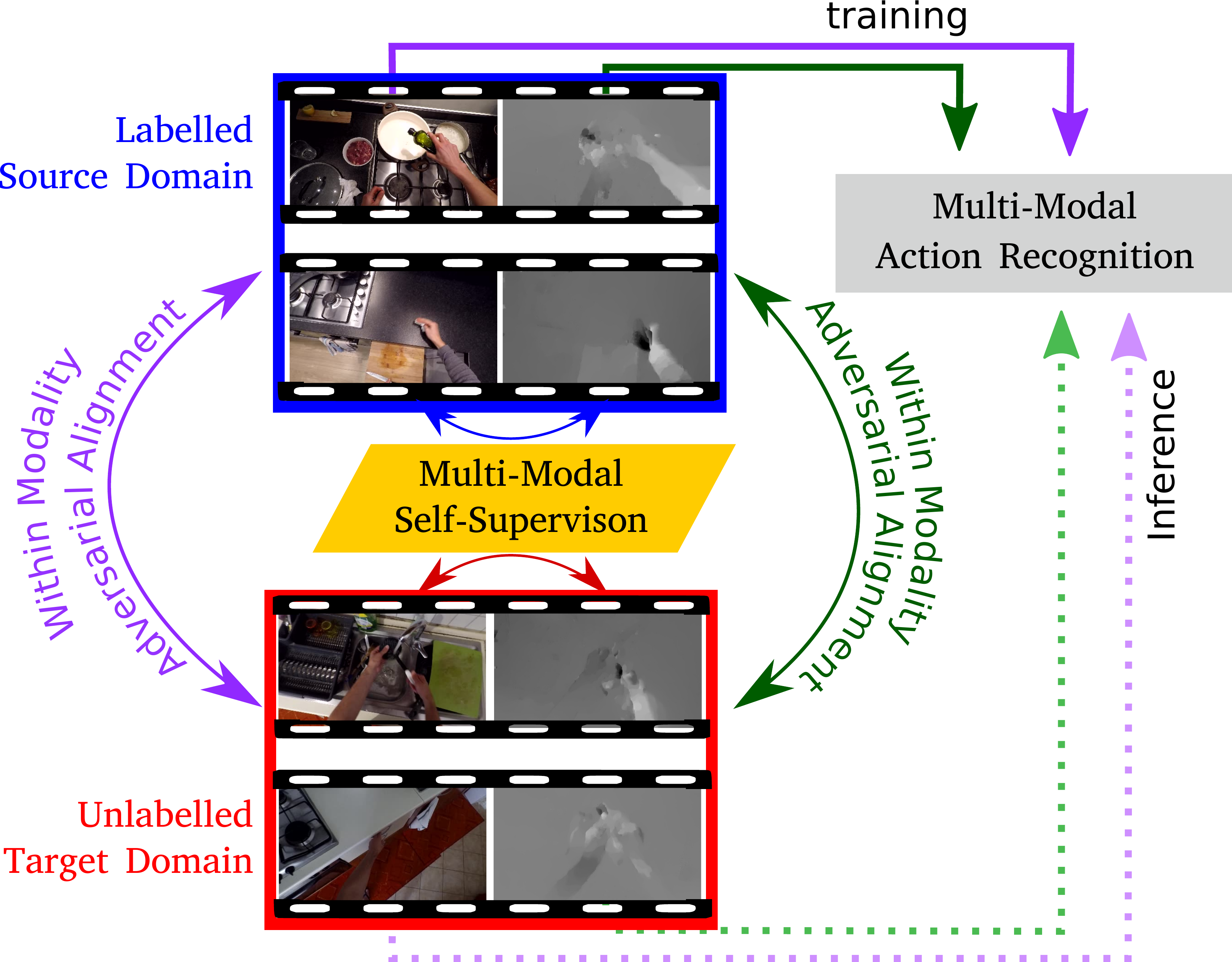}
\caption{Our proposed UDA approach for multi-modal action recognition. Improved target domain performance is achieved via multi-modal self-supervision on source and target domains simultaneously, jointly optimised with multiple domain discriminators, one per-modality.}
\vspace*{-12pt}
\label{fig:motivation}
\end{figure}
Fine-grained action recognition is the problem of recognising actions and interactions such as ``cutting a tomato'' or ``tightening a bolt'' compared to coarse-grained actions such as ``preparing a meal''. This has a wide range of applications in assistive technologies in homes as well as in industry. Supervised approaches rely on collecting a large number of labelled examples to train discriminative models.
However, due to the difficulty in collecting and annotating such fine-grained actions, many datasets collect long untrimmed sequences. These contain several fine-grained actions from a single~\cite{rohrbach15ijcv,Stein2013} or few~\cite{damen2018scaling,sigurdsson2016hollywood} environments. 

\begin{figure}[t]
    \centering
    \includegraphics[width=0.48\textwidth]{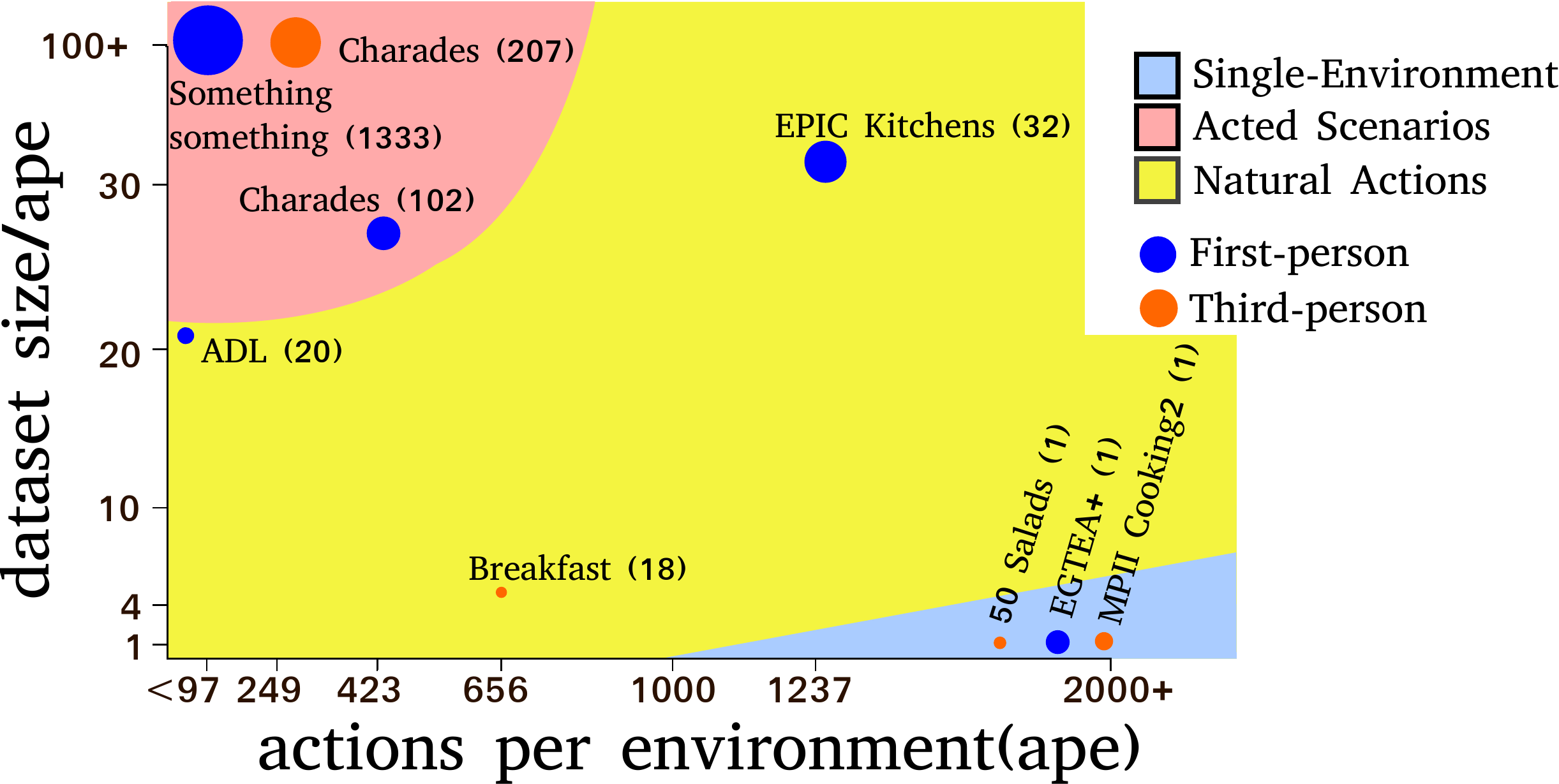}
    \caption{Fine-grained action datasets~\cite{goyal2017something,sigurdsson2016hollywood,sigurdsson2018actor,damen2018scaling,Pirsiavash2012,Kuehne2014,Stein2013,Li_2018_ECCV,Rohrbach2012}, \textit{x-axis:}~number of action segments per environment (ape), \textit{y-axis:} dataset size divided by ape.
   % are able to scale to a large number of environments using scripting actions and crowd-sourced actors. However datasets for fine-grained natural actions collect data from far fewer environments. We evaluate our method on
    EPIC-Kitchens~\cite{damen2018scaling} offers the largest ape relative to its size.}% Notice for those datasets with strong environmental bias, EPIC Kitchens has the largest number of actions.}
    \vspace*{-12pt}
    \label{fig:datasets_compare}
\end{figure}

Figure \ref{fig:datasets_compare} shows the recent surge in large-scale fine-grained action datasets. Two approaches have been attempted to achieve scalability: crowd-sourcing scripted actions~\cite{goyal2017something, sigurdsson2016hollywood,sigurdsson2018actor}, and long-term collections of natural interactions in homes~\cite{damen2018scaling,rohrbach15ijcv,pirsiavash2012detecting}. While the latter offers more realistic videos, many actions are collected in only a few environments.
This leads to learned representations which do not generalise well~\cite{torralba2011unbiased}. %Surprisingly, little research has been conducted to adapt models for fine-grained action recognition to a target environment. 

Transferring a model learned on a labelled source domain to an unlabelled target domain is known as Unsupervised Domain Adaptation (UDA). Recently, significant attention has been given to deep UDA in other vision tasks~\cite{sun2016deep,ghifary2014domain,long2015learning,long2017deep,ganin2016domain,tzeng2017adversarial}. %, including a recent work on multi-modal object recognition~\cite{qi2018unified}. 
%Of these works, adversarial training~\cite{ganin2016domain} encourages learned representations to be indistinguishable between source and target data, while maintaining performance on the source's classification task.
% however only a few Action Recognition works focus on UDA~\cite{jamaldeep,cao2010cross,faraji2011domain,zhu2013enhancing}. %Given the strong environmental bias in Fine-grained Action Recognition, UDA is needed to create more transferable representations without annotation in the target environment that is often difficult or impossible.
However, very few works have attempted deep UDA for video data~\cite{jamaldeep,Chen_2019_ICCV}. Surprisingly, none have tested on videos of fine-grained actions and all these approaches only consider video as images (\ie{} RGB modality). This is in contrast with self-supervised approaches that have successfully utilised multiple modalities within video when labels are not present during training~\cite{arandjelovic2017look}. %Multi-modal self-supervision in video for UDA is currently unexplored.

Up to our knowledge, no prior work has explored the \textbf{multi-modal} nature of video data for UDA in action recognition. %target environments and no work in adapting domain adpatation in video has used self-supervision for domain adaptation. 
We summarise our contributions as follows:

\vspace*{-10pt}

\begin{itemize}[leftmargin=*,itemsep=-2ex,partopsep=1ex,parsep=2ex]
 %   \item We study the robustness of the two most commonly used modalities in action recognition, RGB and Optical Flow, to environmental changes. We show that Flow, which captures motion information, generalises better to target environments.
    \item We show that multi-modal self-supervision, applied to both source and unlabelled target data, can be used for domain adaptation in video.
    \item We propose a multi-modal UDA strategy, which we name MM-SADA, to adapt fine-grained action recognition models to unlabelled target environments, using both adversarial alignment and multi-modal self-supervision. 
    \item We test our approach on three domains from EPIC-Kitchens~\cite{damen2018scaling}, %(Fig~\ref{fig:datasets_compare}),
    trained end-to-end using I3D~\cite{Carreira_2017_CVPR}, and provide the first benchmark of UDA for fine-grained action recognition. Our results show that \mbox{MM-SADA} outperforms source-only generalisation as well as alternative domain adaptation strategies such as batch-based normalisation~\cite{li2018adaptive}, distribution discrepancy minimisation~\cite{long2015learning} and classifier discrepancy~\cite{saito2018maximum}. %Results on 3 different kitchens environments from EPIC-Kitchens show an average of 3.2\% improvement in accuracy.
\end{itemize}

%-------------------------------------------------------------------------
\section{Related Works}
This section discusses related literature starting with general UDA approaches, then supervised and self-supervised learning for action recognition, concluding with works on domain adaptation for action recognition.

\noindent \textbf{Unsupervised Domain Adaptation (UDA) outside of Action Recognition.} UDA has been extensively studied for vision tasks including object recognition~\cite{sun2016deep,ghifary2014domain,long2015learning,long2017deep,ganin2016domain,tzeng2017adversarial}, semantic segmentation~\cite{Zou_2018_ECCV,Huang_2018_ECCV,Zhang_2017_ICCV} and person re-identification~\cite{Sohn_2017_ICCV, Zhong_2018_CVPR,Deng_2018_CVPR}. Typical approaches adapt neural networks by minimising a discrepancy  measure~\cite{sun2016deep, ghifary2014domain}, thus matching mid-level representations of source and target domains.
% over features extracted by convolutional layers to align the statistics of source and target domains. Maximum Mean Discrepancy(MMD) is one of the most popular discrepancy measures for domain adaptation \cite{ghifary2014domain,long2015learning,long2017deep}. 
For example, Maximum Mean Discrepancy~(MMD)~\cite{ghifary2014domain,long2015learning,long2017deep} minimises the distance between the means of the  projected domain distributions in Reproducing Kernel Hilbert Space. More recently, domain adaptation has been influenced by adversarial training~\cite{ganin2016domain,tzeng2017adversarial}. Simultaneously learning a domain discriminator, whilst maximising its loss with respect to the feature extractor, minimises the domain discrepancy between source and target. In~\cite{tzeng2017adversarial}, a GAN-like loss function allows separate weights for source and target domains, while
in~\cite{ganin2016domain} shared weights are used, efficiently removing domain specific information by inverting the gradient produced by the domain discriminator with a Gradient Reversal Layer (GRL). 

Utilising multiple modalities (image and audio) for UDA has been recently investigated for bird image retrieval~\cite{qi2018unified}. Multiple adversarial discriminators are trained on a single modality as well as mid-level fusion and a cross-modality attention is learnt. The work shows the advantages of multi-modal domain adaptation in contrast to single-modality adaptation, though in their work both modalities demonstrate similar robustness to the domain shift.
 
% \noindent \textbf{Self-supervision for UDA and semi-supervised learning.}
% Self-supervision has been exploited for semi-supervision where both unlabelled and labelled data are present~\cite{zhai2019s4l}. With self-supervision tasks, such as predicting angle of rotations, applied to input images, their method could learn from unlabelled data. In our work, we utilise self-supervision for domain adaptation in video where unlabelled data is collected from a different domain to that of labelled examples.

%Domain generalisation techniques have used self-supervision~\cite{}, where multiple labelled source domains are present with the aim to generalise to unseen target domains. By shuffling image patches and predicting the permutation for all domains as an auxiliary task, domain bias could be reduced. 
Very recently, self-supervised learning has been proposed as a domain adaptation approach~\cite{sun2019unsupervised,carlucci2019domain}. 
In~\cite{carlucci2019domain}, it is used as an auxiliary task, by jigsaw-shuffling image patches and predicting their permutations over multiple source domains.
In~\cite{sun2019unsupervised}, self-supervision was shown to replace adversarial training using tasks such as predicting rotation and translation for object recognition. In the same work, 
%For object recognition, self-supervision on tasks, such as rotation and translation, replaces adversarial training. For semantic segmentation, 
self-supervision was shown to benefit adversarial training when jointly trained for semantic segmentation. 
Both works only use a single image. Our work utilises the multiple modalities offered by video, showing that self-supervision can be used to adapt action recognition models to target domains.

\begin{figure*}
	\begin{center}
		\includegraphics[width=1\linewidth]{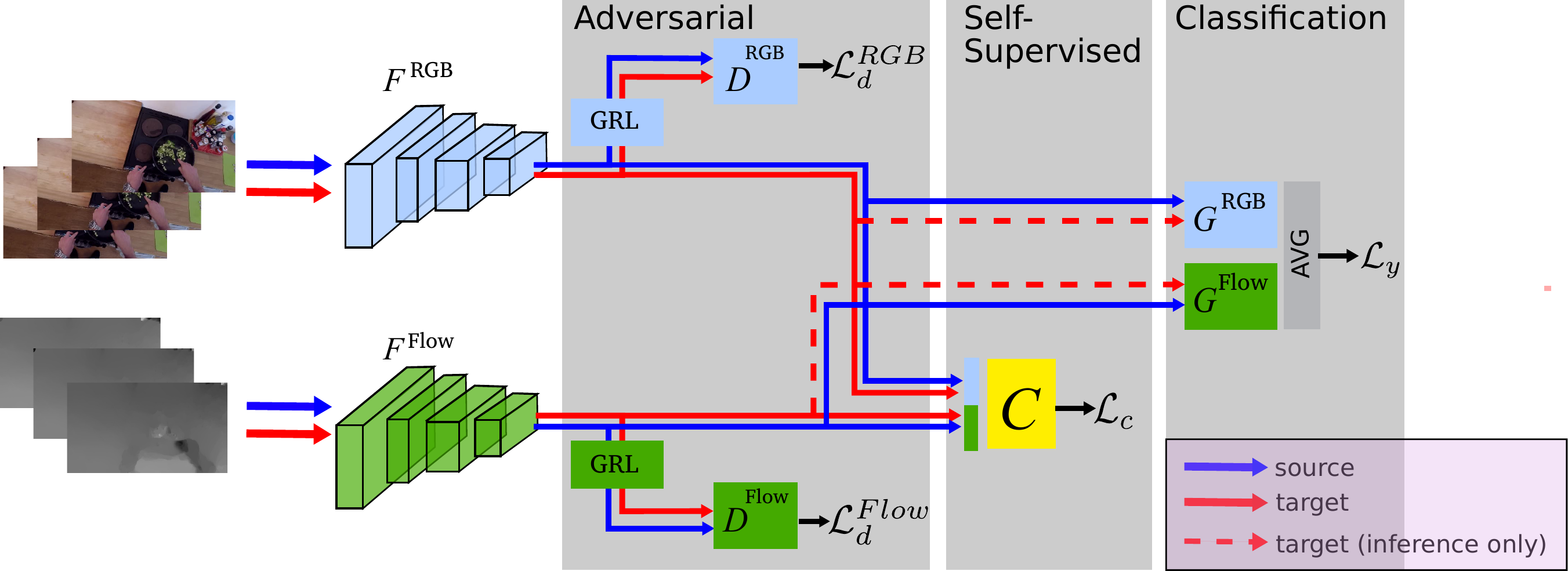}
	\end{center}
	\vspace{-13pt}
	\caption{Proposed architecture: feature extractors {\rgb{$F^{RGB}$}} and {\flow{$F^{Flow}$}} are shared for both {\target{target}} and {\textcolor{blue}{source}} domains. Domain Discriminators, {\rgb{$D^{RGB}$}} and {\flow{$D^{Flow}$}}, are applied to each modality. Self-supervised correspondence of modalities, $C$, is trained from both \textcolor{blue}{source} and \target{unlabelled target} data. Classifiers, {\rgb{$G^{RGB}$}} and {\flow{$G^{Flow}$}} are trained using {\textcolor{blue}{source}} domain examples only from the average pooled classification scores of each modality. During inference, multimodal \target{target} data is classified.} 
	\vspace*{-12pt}
	\label{fig:architecture}
\end{figure*}

\noindent \textbf{Supervised Action Recognition.} 
Convolutional networks are state of the art for action recognition, with the first seminal works using either 3D~\cite{ji20133d} or 2D convolutions~\cite{karpathy2014large}. Both utilise a single modality---appearance information from RGB frames. Simonyan and Zisserman~\cite{NIPS2014_5353} address the lack of motion features captured by these architectures, proposing two-stream late fusion that learns separate features from the Optical Flow and RGB modalities, outperforming single modality approaches.
%In~\cite{Feichtenhofer_2016_CVPR}, mid-level fusion is shown to outperform early fusion for action recognition.

Following architectures have focused on modelling longer temporal structure, through consensus of predictions over time~\cite{TSN2016ECCV,Zhou_2018_ECCV, Lin_2019_ICCV} as well as inflating CNNs to 3D convolutions~\cite{Carreira_2017_CVPR}, all using the two-stream approach of late-fusing RGB and Flow. 
The latest architectures have focused on reducing the high computational cost of 3D convolutions~\cite{Feichtenhofer_2019_ICCV, Jiang_2019_ICCV,ZhaoCVPR2019}, yet still show improvements when reporting results of two-stream fusion~\cite{ZhaoCVPR2019}. 

%Supervised, weakly supervised and unsupervised methods have been proposed for training action recognition networks.
\noindent \textbf{Self-supervision for Action Recognition.}
%Unsupervised methods for action recognition do not require any additional labeled data and directly exploit structure of video to learn representations. Auto-encoders and Restricted Boltzmann's Machines were commonly used to disentangle factors of variation by reconstructing the original input \cite{,,}, with recent research focusing on more relevant supervisory signals capturing semantic information that compliments the recognition task. For action recognition, 
Self-supervision methods learn representations from the temporal~\cite{fernando2017self,wei2018learning}  and multi-modal structure of video~\cite{arandjelovic2017look,korbar2018cooperative}, leveraging pretraining on a large corpus of unlabelled videos. 
Methods exploiting the temporal consistency of video have predicted the order of a sequence of frames~\cite{fernando2017self} or the arrow of time~\cite{wei2018learning}. Alternatively, the correspondence between multiple modalities has been exploited for self-supervision, particularly with audio and RGB~\cite{arandjelovic2017look, korbar2018cooperative, Owens_2018_ECCV}. Works predicted if modalities correspond or are synchronised. We test both approaches for self-supervision in our UDA approach. 

%This work, influenced multi-modal self supervision, proposes to use self-supervision as domain alignment by providing tasks that can be jointly optimised between source and target domains without additional labelled data. By utilising multi-modal self-supervision less robust modalities can benefit from predicting correspondence with the more robust modality. 

 %this work extends this idea utilising self-supervision of multi-modal video to align domains for action recognition.

 %Tzeng et al. \cite{tzeng2017adversarial} extended this framework to allow source and target domains to have separate feature extractor weights, training the network with a similar loss functions to Generative Adversarial Network\cite{goodfellow2014generative}. 
% In this work, we utilise an approach similar to \cite{ganin2016domain}, sharing weights between source and target domains, as the domains in our problem are highly related %as each domain are highly related 
% except for the change in environment.
% so we have chosen to share weights using GRLs to create invariant features. 
%Both GRLs \cite{cao2018partial,zhang2018importance,pei2018multi} and GAN losses \cite{xu2018deep,luo2018taking,luo2017label} have been utilized in recent works. Generative approaches have also been proposed for domain adaptation \cite{sankaranarayanan2018generate}, encouraging the network to produce source-like images for target domain data, however due to large amount of task irreverent information in videos for action recognition, discriminative approaches have been used in this work.

\noindent \textbf{Domain Adaptation for Action Recognition.} 
Of the several domain shifts in action recognition, only one has received significant research attention, that is the problem of cross-viewpoint (or viewpoint-invariant) action recognition~\cite{rahmani2015learning,kong2017deeply,liu2017enhanced,sigurdsson2018actor,li2018unsupervised}.
These works focus on adapting to the geometric transformations of a camera but do little to combat other shifts, like changes in environment. Works utilise supervisory signals such as skeleton or pose~\cite{liu2017enhanced} and corresponding frames from multiple viewpoints~\cite{sigurdsson2018actor,kong2017deeply}.
%to transfer between viewpoints or create viewpoint invariant representations. 
Recent works have used GRLs to create a view-invariant representation~\cite{li2018unsupervised}. Though several modalities (RGB, flow and depth) have been investigated, these were aligned and evaluated independently.

On the contrary, UDA for changes in environment has received limited recent attention.
Before deep-learning, UDA for action recognition used shallow models to align source and target distributions of handcrafted features~\cite{cao2010cross,faraji2011domain,zhu2013enhancing}.
Three recent works attempted deep UDA~\cite{jamaldeep,Chen_2019_ICCV,pan2019adversarial}. These apply GRL adversarial training to C3D~\cite{tran2015learning}, TRN~\cite{Zhou_2018_ECCV} or both~\cite{pan2019adversarial} architectures. Jamal \etal's approach~\cite{jamaldeep} outperforms shallow methods that use subspace alignment. Chen \etal~\cite{Chen_2019_ICCV} show that attending to the temporal dynamics of videos can improve alignment. Pan \etal~\cite{pan2019adversarial} use a cross-domain attention module, to avoid uninformative frames. Two of these works use RGB only~\cite{jamaldeep,Chen_2019_ICCV} while~\cite{pan2019adversarial} reports results on RGB and Flow, however, modalities are aligned independently and only fused during inference. 
The approaches~\cite{pan2019adversarial, Chen_2019_ICCV, jamaldeep} are evaluated on 5-7 pairs of domains from subsets of coarse-grained action recognition and gesture datasets, for example aligning UCF~\cite{reddy2013recognizing} to Olympics~\cite{niebles2010modeling}.
%In~\cite{Chen_2019_ICCV, pan2019adversarial}, their methods are evaluated on 5 pairs of source/target domains, while~\cite{jamaldeep} evaluate their method on 6 pairs of domains from subsets of coarse-grained action datasets such as UCF~\cite{reddy2013recognizing}, Olympics~\cite{niebles2010modeling} and the KMS dataset~\cite{jamaldeep}.%MSR Action II.
%As gameplay - kinetics is larger I've removed statement saying our domains are larger than all.
We evaluate on 6 pairs of domains. Compared to~\cite{jamaldeep}, we use 3.8$\times$ more training and 2$\times$ more testing videos. %Additionally, we focus on fine-grained actions previously unexplored for UDA.

%The EPIC-Kitchens~\cite{damen2018scaling} dataset for fine-grained action recognition has particularly highlighted the need to generalise to novel environments by releasing two distinct test sets - one with seen and another with unseen/novel kitchens.
The EPIC-Kitchens~\cite{damen2018scaling} dataset for fine-grained action recognition released two distinct test sets---one with seen and another with unseen/novel kitchens.
In the 2019 challenges report, all participating entries exhibit a drop in action recognition accuracy of 12-20\% when testing their models on novel environments compared to seen environments~\cite{damen2019technical}. Up to our knowledge, no previous effort applied UDA on this or any fine-grained action dataset.

\noindent \textbf{In this work,} we present the first approach to multi-modal UDA for action recognition, tested on fine-grained actions. We combine adversarial training on multiple modalities with a modality correspondence self-supervision task. This utilises the differing robustness to domain shifts between the modalities. %We show that jointly training for both objectives outperforms 
%single-modality adaptation, as in~\cite{jamaldeep}, as well as using 
%adversarial or self-supervision alignment solely. 
Our method is detailed next.
%In this work, we propose multi-modal domain adaptation for action recognition, recently explored in object detection~\cite{qi2018unified}. We show that multi-modal adaptation using late-fusion outperforms single-modality adaptation, as in~\cite{jamaldeep}. 
%show that adapting the less robust modality (RGB) solely is inferior to our proposed domain adaptation that utilises multiple modalities.
%We showcase our results on fine-grained action recognition~\cite{damen2018scaling}, where the domain shift is primarily due to environmental bias.

%%%%%%%%% Method Section
\section{Proposed Method}

This section outlines our proposed action recognition domain adaptation approach, which we call \textit{Multi-Modal Self-Supervised Adversarial Domain Adaptation (MM-SADA)}. 
In Fig.~\ref{fig:architecture}, we present an overview of MM-SADA, visualised for action recognition using two modalities: \rgb{RGB} and \flow{Optical Flow}. We incorporate a self-supervision alignment classifier, $C$, that determines whether modalities are sampled from the same or different actions to learn modality correspondence.  This takes in the concatenated features from both modalities, without any labels. Learning the correspondence on \source{source} and \target{target} encourages features that generalise to both domains. Aligning the domain statistics is achieved by adversarial training, with a domain discriminator per modality that predicts the domain. A Gradient Reversal layer~(GRL) reverses and backpropagates the gradient to the features. Both alignment techniques are trained on \source{source} and \target{unlabelled target} data whereas the action classifier is only trained with labelled \source{source} data.

We next detail MM-SADA, generalised to any two or more modalities.
We start by revisiting the problem of \textit{domain adaptation} and outlining multi-stream late fusion, then we describe our adaptation approach.

\subsection{Unsupervised Domain Adaptation (UDA)}
\vspace*{-3pt}
A domain is a distribution over the input population \textbf{X} and the corresponding label space \textbf{Y}. The aim of supervised learning, given labelled samples~$\{(x, y)\}$, is to find a representation, $G(\cdot)$, over some learnt features, $F(\cdot)$, that minimises the empirical risk, $\mathop{E_\mathbf{S}}[\mathcal{L}_y(G(F(x)),y)]$. The empirical risk is optimised over the labelled source domain, $\mathbf{S}=\{X^s,Y^s,\mathcal{D}^s\}$, where $\mathcal{D}^s$ is a distribution of source domain samples. The goal of domain adaptation is to minimise the risk on a target domain, $\mathbf{T}=\{X^t,Y^t,\mathcal{D}^t\}$, where the distributions in the source and target domains are distinct, ${\mathcal{D}^s \neq \mathcal{D}^t}$. In~UDA, the label space $Y^t$ is unknown, thus methods minimise both the source risk and the distribution discrepancy between the source and target domains~\cite{NIPS2006_2983}.

\subsection{Multi-modal Action Recognition}
\vspace*{-3pt}
When the input is multi-modal,  \ie{}~${X = (X^1, \cdots, X^M)}$ where $X^m$ is the $m^{th}$ modality of the input, fusion of modalities can be employed. Most commonly, late fusion is implemented, where we sum prediction scores from modalities and backpropagate the error to all modalities, \ie{}:

\vspace*{-16pt}
\begin{multline}
\mathcal{L}_y = \sum_{x \in \{\textbf{S}\}}  -y \log P(x) \\ 
\text{where: } P(x) = \sigma\big(\sum^M_{m=1}G^m(F^m(x^m))\big)
\label{eq:mmClassify}
\end{multline}
where $G^m$ is the modality's task classifier, and $F^m$ is the modality's learnt feature extractor. The consensus of modality classifiers is trained by a cross entropy loss, $\mathcal{L}_y$, between the task label, $y$, and the prediction, $P(x)$. $\sigma$ is defined as the softmax function. Training for classification expects the presence of labels and thus can only be applied to the labelled source input.

\subsection{Within-Modal Adversarial Alignment}
\vspace*{-3pt}
Both generative and discriminative adversarial approaches have been proposed for bridging the distribution discrepancy between source and target domains. Discriminative approaches are most appropriate with high-dimensional input data present in video. Generative adversarial requires a huge amount of training data and temporal dynamics are often difficult to reconstruct. %only a small amount of information in videos is discriminative to the action recognition task.
Discriminative methods train a discriminator, $D(\cdot)$, to predict the domain of an input (\ie{} source or target), from the learnt features, $F(\cdot)$. By maximising the discriminator loss, the network learns a feature representation that is invariant to both domains.
 
For aligning multi-modal video data, we propose using a domain discriminator per modality that penalises domain specific features from each modality's stream. 
Aligning modalities separately avoids the easier solution of the network focusing only on the less robust modality in classifying the domain.  
Each separate domain discriminator, $D^m$, is thus used to train the modality's feature representation~$F^m$.  Given a binary domain label, $d$, indicating if an example $x \in \textbf{S}$ or $x \in \textbf{T}$, the domain discriminator, for modality $m$, is defined as,

\vspace*{-12pt}
\begin{multline}
    \mathcal{L}^m_{d} = \sum_{x \in \{\textbf{S}, \textbf{T}\}} -d \log(D^m(F^m(x))) - \\
    (1-d) \log(1-D^m(F^m(x)))
    \label{eq:adv}
\end{multline}

\subsection{Multi-Modal Self-Supervised Alignment}
\vspace*{-4pt}
Prior approaches to domain adaptation have mostly focused on images and thus have not explored the multi-modal nature of the input data.
Videos are multi-modal, where corresponding modalities are present in both source and target.
We thus propose a multi-modal self-supervised task to align domains.
Multi-modal self-supervision has been successfully exploited as a pretraining strategy~\cite{arandjelovic2017look,arandjelovic2018objects}. However, we show that self-supervision for both source and target domains can also align domains.
% We thus propose to add self-supervised tasks along with DADA training.
% The pre-text task of multi-modal alignment has been successfully exploited as a pretraining strategy \cite{arandjelovic2017look,arandjelovic2018objects}.
 %We propose approaches have been used for unlabelled data, where the task of alignment modalities requires no further labels. }
%We propose to learn multiple supervisory tasks to align domains, finding a common representation to fit both source and target domains. However in Unsupervised Domain Adaptation, no labels can be collected in the target domain. Therefore any supervisory tasks in addition to action classification must be self-supervised, learning only from the data-stream. This method utilises the synchrony of multiple modalities as a self-supervision techniques to learn representations that generalise to the target domain.
%%% Discus the use of synchrony detection as a side task
%The synchronisation between modalities has been exploited successfully as a pretraining task \cite{arandjelovic2017look,arandjelovic2018objects}, requiring the network learn representations useful for action recognition. 
%We propose a classifier that predicts whether modalities are corresponding, shared between source and target data, as a domain alignment approach.
%to predict whether modalities are synchronised as a domain alignment approach. 
%By jointly optimising DADA with this self-supervised task, features that generalise to both source and target domains are learnt. 

%% Network Archetecture.

We learn the temporal correspondence between modalities as a self-supervised binary classification task. For positive examples, indicating that modalities correspond, we sample modalities from the same action. These could be from the same time, or different times within the same action. For negative examples, each modality is sampled from a different action. The network is thus trained to determine if the modalities correspond. This is optimised over both domains.
A self-supervised correspondence classifier head, $C$, is used to predict if modalities correspond. This shares the same modality feature extractors, $F^m$, as the action classifier. It is important that $C$ is as shallow as possible so that most of the self-supervised representation is learned in the feature extractors. Given a binary label defining if modalities correspond, $c$, for each input, $x$, and concatenated features of the multiple modalities, we calculate the multi-modal self-supervision loss as follows:
\begin{equation}
    \mathcal{L}_c = \sum_{x \in \{\textbf{S,T}\}}  -c \log C(F^0(x), ..., F^M(x))%\\
    %\text{where: } f^m = F^m(x)
    \label{eq:mmSelf}
\end{equation}

\subsection{Proposed MM-SADA}%Multi-modal Self-Supervised Action Domain Adaptation}
We define the Mutli-Modal Self-Supervised Adversarial Domain Adaptation (MM-SADA) approach as follows. The classification loss, $\mathcal{L}_y$, is jointly optimised with the adversarial and self-supervised alignment losses. The within-modal adversarial alignment is weighted by $\lambda_d$, and the multi-modal self-supervised alignment is weighted by $\lambda_c$. Optimising both alignment strategies achieves benefits in matching source and target statistics and learning cross-modal relationships transferable to the target domain.
%Finally the Mutli-modal Self-Supervised Adversarial Domain Adaptation (MM-SADA) is jointly optimised as the sum of the various classification and alignment losses.
\begin{equation}
\mathcal{L} = \mathcal{L}_y + \lambda_d \sum_m \mathcal{L}_d^m + \lambda_c \mathcal{L}_c
\label{eq:mmLoss}
\end{equation}
Note that the first loss $\mathcal{L}_y$ is only optimised for labelled source data, while the alignment losses $\forall m:\mathcal{L}_d^m$ and $\mathcal{L}_c$ are optimised for both unlabelled source and target data.

%-----------------------------------------------------------------------\subsection{Experimental Setup}
\section{Experiments and Results}

This section first discusses the dataset, architecture, and implementation details in Sec.~\ref{sec:implementation}. We compare against baseline methods noted in Sec.~\ref{sec:baselines}. Results are presented in Sec.~\ref{sec:results}, followed by an ablation study of the method's components in Sec.~\ref{sec:ablation} and qualitative results including feature space visualisations in Sec.~\ref{sec:qualitative}.

\subsection{Implementation Details}
\label{sec:implementation}
\begin{figure}
    \hspace*{\fill}%
    \includegraphics[width=0.98\linewidth]{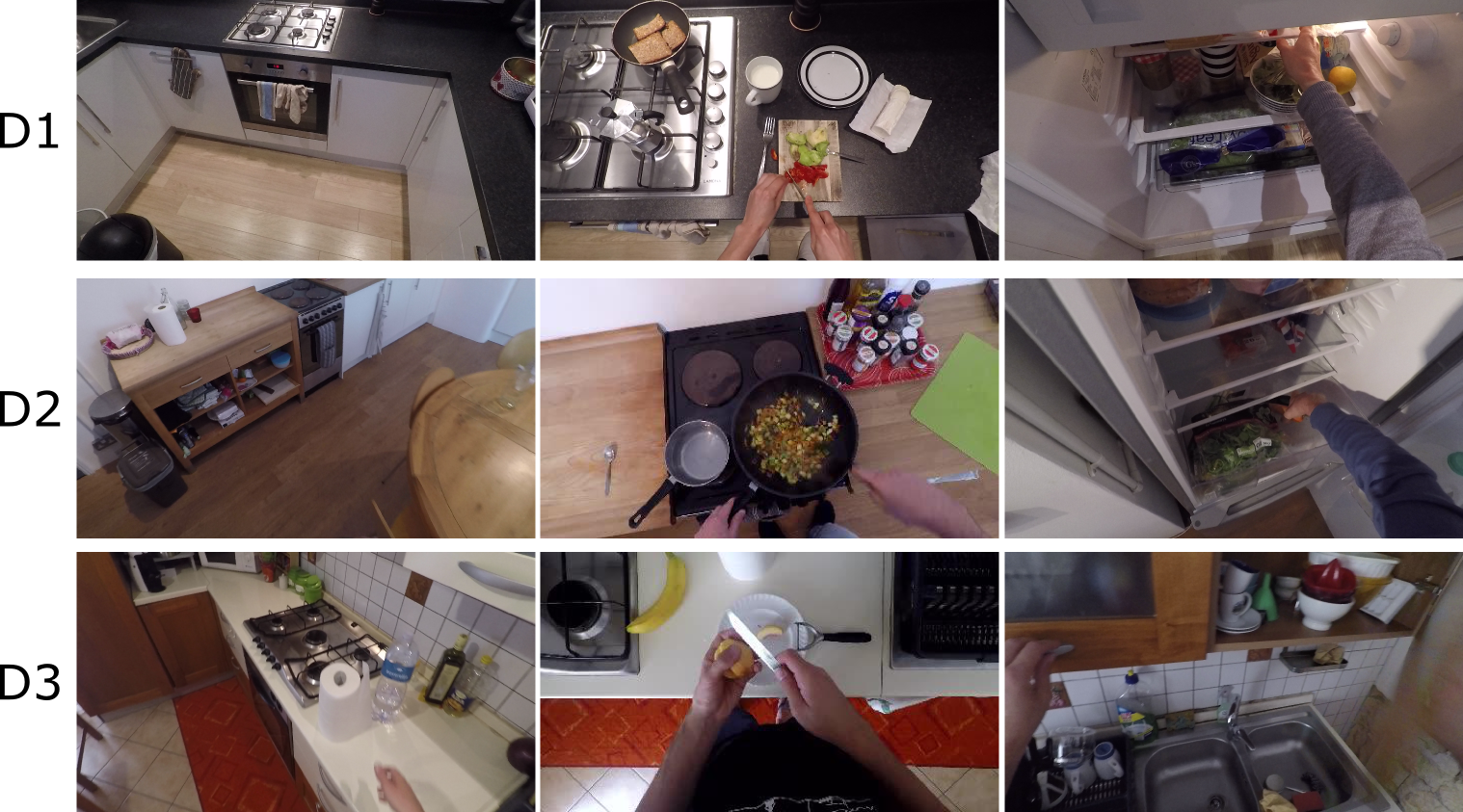}
    \includegraphics[width=0.98\linewidth]{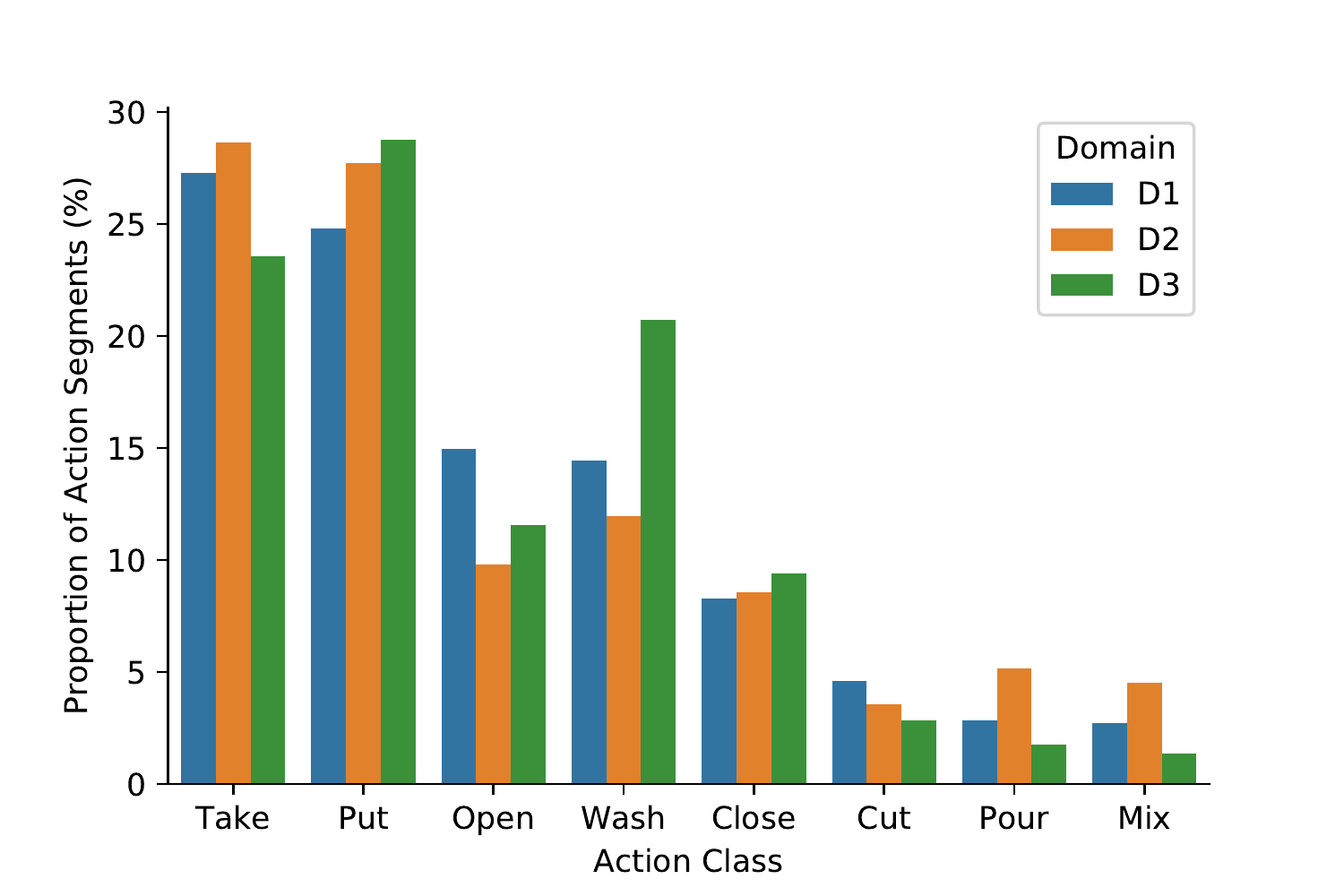}
       % \vspace{20pt}
        %\resizebox{0.8\linewidth}{!}{
        \centering
        \begin{tabular}{l rrr}
            \toprule
             Domain   & D1 & D2 & D3\\
             \midrule
             Ref. EPIC Kitchen  &   P08&  P01      & P22\\
             Training Action Segments   &   1543       &  2495    & 3897\\
             Test Action Segments      &   435  &  750     & 974\\
             \bottomrule
        \end{tabular}%}
    \caption{
    \textbf{Top:} Three kitchens from EPIC-Kitchens selected as domains to evaluate our method
    \textbf{Middle:} Class distribution per domain, for the 8 classes in legend. \textbf{    Bottom:} Number of action segments per domain.}
    \vspace*{-10pt}
    \label{fig:class_proportions}
\end{figure}

\vspace*{-6pt}
\myparagraph{Dataset}
% 8 most common verbs - 78% of EPIC Kitchne
% 3 largest kitchens - 32% of EPIC Kitchnes
Our previous work, EPIC Kitchens~\cite{damen2018scaling}, offers a unique opportunity to test domain adaptation for fine-grained action recognition, as it is recorded in 32 environments.
Similar to previous works for action recognition~\cite{ganin2016domain,jamaldeep}, we evaluate on pairs of domains. We select the three largest kitchens, in number of training action instances, to form our domains. These are P01, P22, P08, which we refer to as D1, D2 and D3, respectively~(Fig.~\ref{fig:class_proportions}).

We analyse the performance for the 8 largest action classes: (`put', `take', `open', `close', `wash', `cut', `mix', and `pour'), which form 80\% of the training action segments for these domains. This ensures sufficient examples per domain and class, without balancing the training set. The label imbalance of these 8 classes is depicted in Fig.~\ref{fig:class_proportions} (middle) which also shows the differing distribution of classes between the domains. Most domain adaptation works evaluate on balanced datasets~\cite{10.1007/978-3-642-15561-1_16,ganin2016domain,gong2012geodesic} with few using imbalanced datasets~\cite{venkateswara2017deep}. EPIC-Kitchens has a large class imbalance offering additional challenges for domain adaptation.
The number of action segments in each domain are specified in Fig.~\ref{fig:class_proportions}~(bottom), where a segment is a labeled start/end time, with an action label. 
\begin{table*}[]
    \centering
    %\resizebox{0.95\textwidth}{!}{%
\begin{tabular}{l l l l l l l l l} 
\toprule
&  D2$\rightarrow$ D1 &  D3$\rightarrow$ D1 &  D1$\rightarrow$ D2 &  D3$\rightarrow$ D2 &  D1$\rightarrow$ D3 &  D2$\rightarrow$ D3 &     Mean \\
\midrule
MM Source-only & 42.5 & 44.3 & 42.0 & \textbf{56.3} &  41.2 & 46.5& 45.5 \\
AdaBN~\cite{li2018adaptive}         &              44.6   &             47.8    &     47.0            &         54.7        &             40.3      &       48.8          &   47.2  \\
MMD~\cite{long2015learning} & 43.1 & 48.3 & 46.6 & 55.2 & 39.2 & 48.5 & 46.8\\
MCD~\cite{saito2018maximum} & 42.1 & 47.9 & 46.5 & 52.7 & 43.5 & 51.0 & 47.3\\
\midrule
MM-SADA &                \textbf{48.2} \textcolor{ForestGreen}{$\blacktriangle \text{+}5.7$} &                \textbf{50.9}\textcolor{ForestGreen}{$\blacktriangle \text{+}6.6$} &               \textbf{49.5}\textcolor{ForestGreen}{$\blacktriangle \text{+}7.5$} &               56.1\textcolor{BrickRed}{$\blacktriangledown \text{-}0.2$} &                \textbf{44.1}\textcolor{ForestGreen}{$\blacktriangle \text{+}2.9$} &                \textbf{52.7} \textcolor{ForestGreen}{$\blacktriangle \text{+}6.3$} &  \textbf{50.3} \textcolor{ForestGreen}{$\blacktriangle \text{+}4.8$} \\
\midrule
\rowcolor{lightgray}
Supervised target & 62.8 & 62.8 & 71.7 & 71.7 & 74.0 & 74.0 & 69.5 \\
\bottomrule
\end{tabular}
%}
\vspace*{-6pt}
    \caption{Top-1 Accuracy on the target domain, for our proposed MM-SADA, compared to different alignment approaches. On average, we outperform the source-only performance by 4.8\%.}
    \vspace*{-8pt}
    \label{tab:Compare}
\end{table*}
\begin{figure*}[th]
    \begin{subfigure}[]{0.32\textwidth}
        \includegraphics[width=\textwidth]{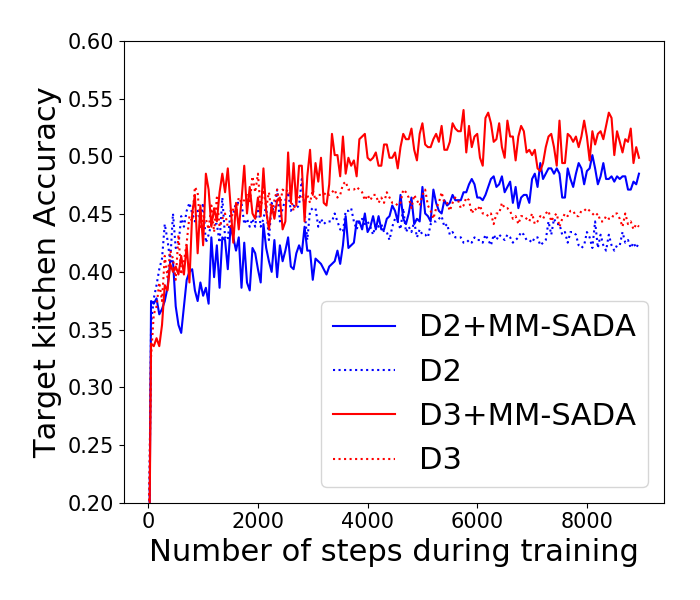}
        \caption{Target D1}
    \end{subfigure}
        \begin{subfigure}[]{0.32\textwidth}
        \includegraphics[width=\textwidth]{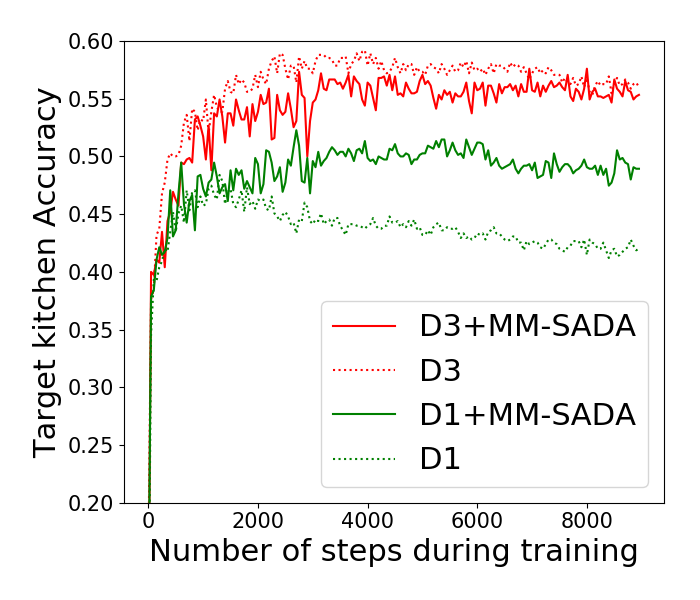}
        \caption{Target D2}
    \end{subfigure}
        \begin{subfigure}[]{0.32\textwidth}
        \includegraphics[width=\textwidth]{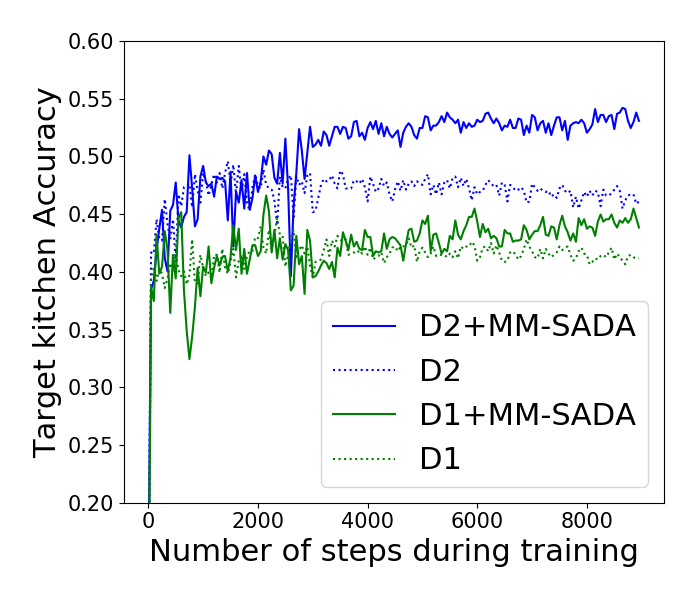}
        \caption{Target D3}
    \end{subfigure}
    \vspace*{-8pt}
    \caption{Accuracy on target during training epochs. Solid line is MM-SADA and dotted line is source-only performance.}
    \label{fig:training}
\end{figure*}

\myparagraph{Architecture}
We train all our models end-to-end.
 We use the inflated 3D convolutional architecture  (I3D)~\cite{Carreira_2017_CVPR} as our backbone for feature extraction, one per modality ($F^m$). In this work, $F$ convolves over a temporal window of 16~frames. 
In training, a single temporal window is randomly sampled from within the action segment each iteration.
In testing, as in~\cite{TSN2016ECCV}, we use an average over 5 temporal windows,  
equidistant within the segment. 
We use the RGB and Optical Flow frames provided publicly~\cite{damen2018scaling}. 
%Only if the segment is shorter than temporal window, neighbouring frames were taken with the window centered on the middle of the segment. 
The output of $F$ is the result of the final average pooling layer of I3D, with 1024 dimensions. $G$ is a single fully connected layer with softmax activation to predict class labels. Each domain discriminator $D^m$ is composed of 2 fully connected layers with a hidden layer of 100 dimensions and a ReLU activation function. A dropout rate of 0.5 was used on the output of $F$ and $1e-7$ weight decay for all parameters. Batch normalisation layers are used in $F^m$ and are updated with target statistics for testing, as in AdaBN~\cite{li2018adaptive}. We apply random crops, scale jitters and horizontal flips for data augmentation as in~\cite{TSN2016ECCV}. 
%Image widths and heights are scaled by $\{1, 0.875, 0.75\}$ during scale jittering and are resized to 224.
During testing only center crops are used.
The self-supervised correspondence function $C$ (Eq.~\ref{eq:mmSelf}) is implemented as 2 fully connected layers of 100 dimensions and a ReLU activation function. The features from both modalities are concatenated along the channel dimension as input to $C$.

\myparagraph{Training and Hyper-parameter Choice}
We train using the Adam optimiser~\cite{adam_citation} in two stages. First the network is trained with only the classification and self supervision losses $\mathcal{L}_{y}+\lambda_c \mathcal{L}_c$ at a learning rate of $1e-2$ for 3K iterations. Then, the overall loss function~(Eq.~\ref{eq:mmLoss}) is optimised, applying the domain adversarial losses $\mathcal{L}^m_{d}$, and reducing the learning rate to $2e-4$ for a further 6K steps. The self-supervision hyper-parameter, $\lambda_{c}=5$ was chosen by observing the performance on the labelled \textbf{source domain} only, \ie{} this has not been optimised for the target domain. Note that while training with self-supervision, half the batch contains corresponding modalities and the other non-corresponding modalities. Only source examples with corresponding modalities are used to train for action classification. The domain adversarial hyper-parameter, $\lambda_d=1$, was chosen arbitrarily; we show that the results are robust to some variations in this hyper-parameter in an ablation study.
 Batch size was set to 128, split equally for source and target samples.
 On average, training takes 9 hours on an NVIDIA DGX-1 with 8 V100 GPUs. %On average, training takes 9 hours.

% \begin{figure*}[th]
%     \begin{subfigure}[]{0.32\textwidth}
%         \includegraphics[width=\textwidth]{images/TrainingCurvesP08.png}
%         \caption{Target D1}
%     \end{subfigure}
%         \begin{subfigure}[]{0.32\textwidth}
%         \includegraphics[width=\textwidth]{images/TrainingCurvesP01.png}
%         \caption{Target D2}
%     \end{subfigure}
%         \begin{subfigure}[]{0.32\textwidth}
%         \includegraphics[width=\textwidth]{images/TrainingCurvesP22.png}
%         \caption{Target D3}
%     \end{subfigure}
%     \vspace*{-8pt}
%     \caption{Accuracy on target during training epochs. Solid line is MM-SADA and dotted line is source-only performance.}
%     \label{fig:training}
% \end{figure*}

\subsection{Baselines}
\label{sec:baselines}
\vspace*{-6pt}
For all results, we report the top-1 target accuracy averaged over the last 9 epochs of training, for robustness. %We also show that the performance is consistent over training epochs, through accuracy curves.
We first evaluate the impact of domain shift between source and target by testing using a multi-modal source-only model (MM source-only), trained with no access to unlabelled target data. 
Additionally, we compare to 3 baselines for unsupervised domain adaptation as follows:

\vspace*{-6pt}
\begin{itemize}[label=--,leftmargin=*]
    \item \textit{AdaBN}~\cite{li2018adaptive}: Batch Normalisation layers are updated with target domain statistics.
    
\vspace*{-2pt}
    \item \textit{Maximum Mean Discrepancy (MMD)}: The multiple kernel implementation of the commonly used domain discrepancy measure MMD is used as a baseline~\cite{long2015learning}. This directly replaces the adversarial alignment with separate discrepancy measures applied to individual modalities.

\vspace*{-2pt}
    \item \textit{Maximum Classifier Discrepancy (MCD)}~\cite{saito2018maximum}: Alignment through classifier disagreement is used. We use two multi-modal classification heads as separate classifiers. The classifiers are trained to maximise prediction disagreement on the target domain, implemented as L1 loss, finding examples out of support from the source domain. We use a GRL to optimise the feature extractors. %, measure the disagreement between their predictions. \dimaN{This is not clear, what do you backpropagate? L2 Loss between the predictions?} 
\end{itemize}

\vspace*{-6pt}

Additionally, as an upper limit, we also report the supervised target domain results. This is a model trained on labelled target data and only offers an understanding of the upper limit for these domains. We highlight these results in the table to avoid confusion.

%-------------------------------------------------------------------------
\subsection{Results}
\label{sec:results}
\vspace*{-6pt}

\begin{table*}[]
    \centering
    \resizebox{\textwidth}{!}{%
        \begin{tabular}{ l S S c c c c c c l } 
        \toprule
        &$\lambda_{d}$ &$\lambda_c$&  D2$\rightarrow$ D1 &  D3$\rightarrow$ D1 &  D1$\rightarrow$ D2 &  D3$\rightarrow$ D2 &  D1$\rightarrow$ D3 &  D2$\rightarrow$ D3 &     Mean \\
        \midrule
        Source-only &0 &0 & 42.5 & 44.3 & 42.0 & 56.3 &  41.2 & 46.5& 45.5 \\
        \midrule
        MM-SADA ({\small{Self-Supervised only}}) &0 &5      &              41.8   &              49.7   &              47.7   &      \textbf{57.4}            &             40.3    &             50.6    &   47.9\textcolor{ForestGreen}{$\blacktriangle \text{+}2.4$} \\
        \midrule
        MM-SADA ({\small{Adversarial only}}) &1 &0       &               46.5 &               51.0 &                50.0 &                53.7 &                     43.5 &                51.5 &  49.4\textcolor{ForestGreen}{$\blacktriangle \text{+}3.9$} \\
        MM-SADA ({\small{Adversarial only}}) & 0.5 & 0 & 46.9 & 50.2 & 50.2 & 53.6 & \textbf{44.7} & 50.8 & 49.4\textcolor{ForestGreen}{$\blacktriangle \text{+}3.9$}\\
        \midrule
        MM-SADA & 0.5 & 5 & 45.8 & \textbf{52.1} & \textbf{50.4} & 56.9 & 43.5 & 51.9 & 50.1\textcolor{ForestGreen}{$\blacktriangle \text{+}4.6$}\\
        MM-SADA &1 &5 &                \textbf{48.2} &                50.9 &               49.5 &               56.1 &                44.2 &                \textbf{52.7} &  \textbf{50.3}\textcolor{ForestGreen}{$\blacktriangle\text{+}4.8$} \\
        \bottomrule
        \hline
    \end{tabular}
    }
     \caption{Ablation of our method, showing the contribution of the various loss functions (Eq~\ref{eq:mmLoss}). When $\lambda_d = 0$, modality adversarial is not utilised. When $\lambda_c=0$, self-supervision is not utilised.}
    \label{tab:Ablation}
    \end{table*}
    \begin{table*}[]
    \centering
    \resizebox{0.82\textwidth}{!}{%
    \begin{tabular}{ l c c c c c c l } 
        \toprule
          & D2$\rightarrow$ D1 &  D3$\rightarrow$ D1 &  D1$\rightarrow$ D2 &  D3$\rightarrow$ D2 &  D1$\rightarrow$ D3 &  D2$\rightarrow$ D3 &     Mean \\
        % Trained from scratch
        % RGB source-only & 1 & 0 & 35.275 & 36.169 & 36.652& 46.119& 35.467 & 37.041 & 37.787\\
        % \hline
        % RGB ADA & 1 & 0 & 41.686 & 38.902 & 43.881 & 46.563 & 37.292 & 44.376 & 42.117\\
        % \hline\hline
        % Flow source-only & 1 & 0 & 46.207 & 44.879 & 52.311 & 59.289 & 42.631 & 48.255 & 48.951\\
        % \hline
        % Flow ADA & & & & & & & & &\\
        % \hline
        % \hline
        
        % Evaluated from the fusion model
         \midrule
         RGB source-only & 37.0 & 36.3 & 36.1 & 44.8 & 36.6 & 33.6 & 37.4\\
         RGB (Adversarial-only) & 37.8 & 41.1 & 45.7 & 45.1 & 38.1 & 41.2 & 41.5\\
         RGB (MM-SADA) & 41.7 & 42.1 & 45.0 & 48.4 & 39.7 & 46.1 & 43.9\\
         \bottomrule
         \toprule
         Flow source-only & 44.6 & 44.4 & 52.2 & 54.0 & 41.1 & 50.0 & 47.7\\
         Flow (Adversarial-only) & 45.5 & 46.8 & 51.1 & 54.6 & 44.2 & 47.1 & 48.2\\
         Flow (MM-SADA) & 45.0 & 45.7 & 49.0 & 58.9 & 44.8 & 52.1 & 49.3\\
        \bottomrule
        \hline
    \end{tabular}
    }

    \caption{Ablation of our method on individual modalities, reporting predictions from each modality stream, before late fusion. Note that we still use both modalities for self-supervision. MM-SADA provides improvements for both modalities.}
    \label{tab:ModalityAblation}
    \vspace*{-8pt}

%      \resizebox{\textwidth}{!}{%
%         \begin{tabular}{ r c c c c c c l } 
%         \toprule
%         &  D2$\rightarrow$ D1 &  D3$\rightarrow$ D1 &  D1$\rightarrow$ D2 &  D3$\rightarrow$ D2 &  D1$\rightarrow$ D3 &  D2$\rightarrow$ D3 &     Mean \\
%         \midrule
%         source-only & 42.5 & 44.3 & 42.0 & 56.3 &  41.2 & 46.5& 45.5 \\
%         \midrule
%         source-only (eval. rgb) & 37.0 & 36.3 & 36.1 & 44.8 & 36.6 & 33.6 & 37.4\\
%         MM-SADA (eval. rgb) & 41.7 & 42.1 & 45.0 & 48.4 & 39.7 & 46.1 & 43.9\\
%         \midrule
%         source-only (eval. flow) & 44.6 & 44.4 & 52.2 & 54.0 & 41.1 & 50.0 & 47.7\\
%         MM-SADA (eval. flow) & 45.0 & 45.7 & 49.0 & \textbf{58.9} & \textbf{44.8} & 52.1 & 49.3\\
%         \midrule
%         MM-SADA &                \textbf{48.2} &                \textbf{50.9} &               \textbf{49.5} &               56.1 &                44.2 &                \textbf{52.7} &  \textbf{50.3}\\
% \bottomrule
% \hline
% \end{tabular}
% }
%     \caption{Comparison of the individual performances of modalities for MM-SADA. Alignment of MM-SADA improves the performance for both modalities but RGB benefits most.}
%     \label{tab:Modalities}  
        \end{table*}
    \begin{table*}[]
    \centering
    \resizebox{0.78\textwidth}{!}{%
\begin{tabular}{l c c c c c c c c } 
\toprule
Self-Supervision &  D2$\rightarrow$ D1 &  D3$\rightarrow$ D1 &  D1$\rightarrow$ D2 &  D3$\rightarrow$ D2 &  D1$\rightarrow$ D3 &  D2$\rightarrow$ D3 &     \textbf{Mean} \\
\hline
Sync.      &       44.2          &   50.2              &     48.0            &        54.6          &      41.0           &             49.4    &  \textbf{47.9}  \\
Seg. Corr.    &              41.8   &              49.7   &              47.7   &      57.4            &             40.3    &             50.6    &   \textbf{47.9} \\
\bottomrule
\end{tabular}}
\caption{Comparision of two self-supervision tasks for modality correspondence: determining modality synchrony vs. determining whether modality samples come from the same segment. The two approaches perform comparably on average.}
%Ablation of MM-SADA over two multi-modal self-supervised tasks: determining modality synchrony vs. determining whether modality samples come from the same segment. The two approaches perform comparably.}

    \label{tab:AblatSynch}
\end{table*}

First we compare our proposed method MM-SADA to the various domain alignment techniques in Table \ref{tab:Compare}. We show that our method outperforms batch-based~\cite{li2018adaptive} (by~3.1\%), classifier discrepancy~\cite{saito2018maximum} (by 3\%) and discrepancy minimisation alignment~\cite{long2015learning} (by 3.5\%) methods. The improvement is consistent for all pairs of domains. 
Additionally, it significantly improves on the source-only baseline by up to 7.5\% in 5 out of 6 cases. For a single case, $D3\rightarrow D2$, all baselines under-perform compared to source-only. Ours has a slight drop~(-0.2\%) but outperforms other alignment approaches. We will revisit this case in the ablation study. % ill-suited for this kitchen pair. Our method has the lowest drop in performance due to the addition of self-supervision, we show that self-supervision alone improves performance for this failure case in the ablation study.  

\begin{figure}
    \includegraphics[width=\linewidth]{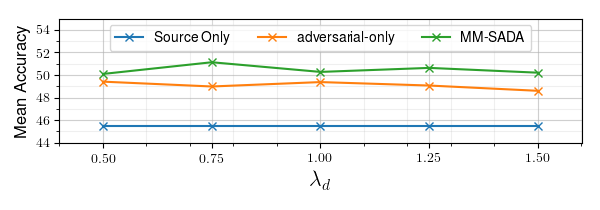}
    \vspace*{-25pt}
    \caption{Robustness of the average top-1 accuracy over all pairs of domains for various $\lambda_d$ on the target domain.}
    \label{fig:robustness}
\end{figure}

Figure~\ref{fig:training} shows the top-1 accuracy on the target during training (solid lines) vs source-only training without domain adaptation (dotted lines). Training without adaptation has consistently lower accuracy, except for our failure case $D3\rightarrow D2$, showing the stability and robustness of our method during training, with minimal fluctuations due to stochastic optimisation on batches. This is essential for UDA as no target labels can be used for early stopping.

\begin{figure*}[t]

    \centering
    % \begin{subfigure}[]{0.9\textwidth}
    %     \includegraphics[width=\textwidth]{images/Correct_Examples_P01_P08.pdf}
    % \end{subfigure}
    % \begin{subfigure}[]{0.9\textwidth}
    %         \includegraphics[width=\textwidth]{images/Correct_Examples_P01_P22.pdf}
    % \end{subfigure}
    % \caption{Qualitative results for MM-SADA and source-only, showing success and failure cases.}
    % \label{fig:qualitative}
    \vspace{6pt}
\begin{subfigure}[]{0.48\textwidth}
        \includegraphics[width=\textwidth]{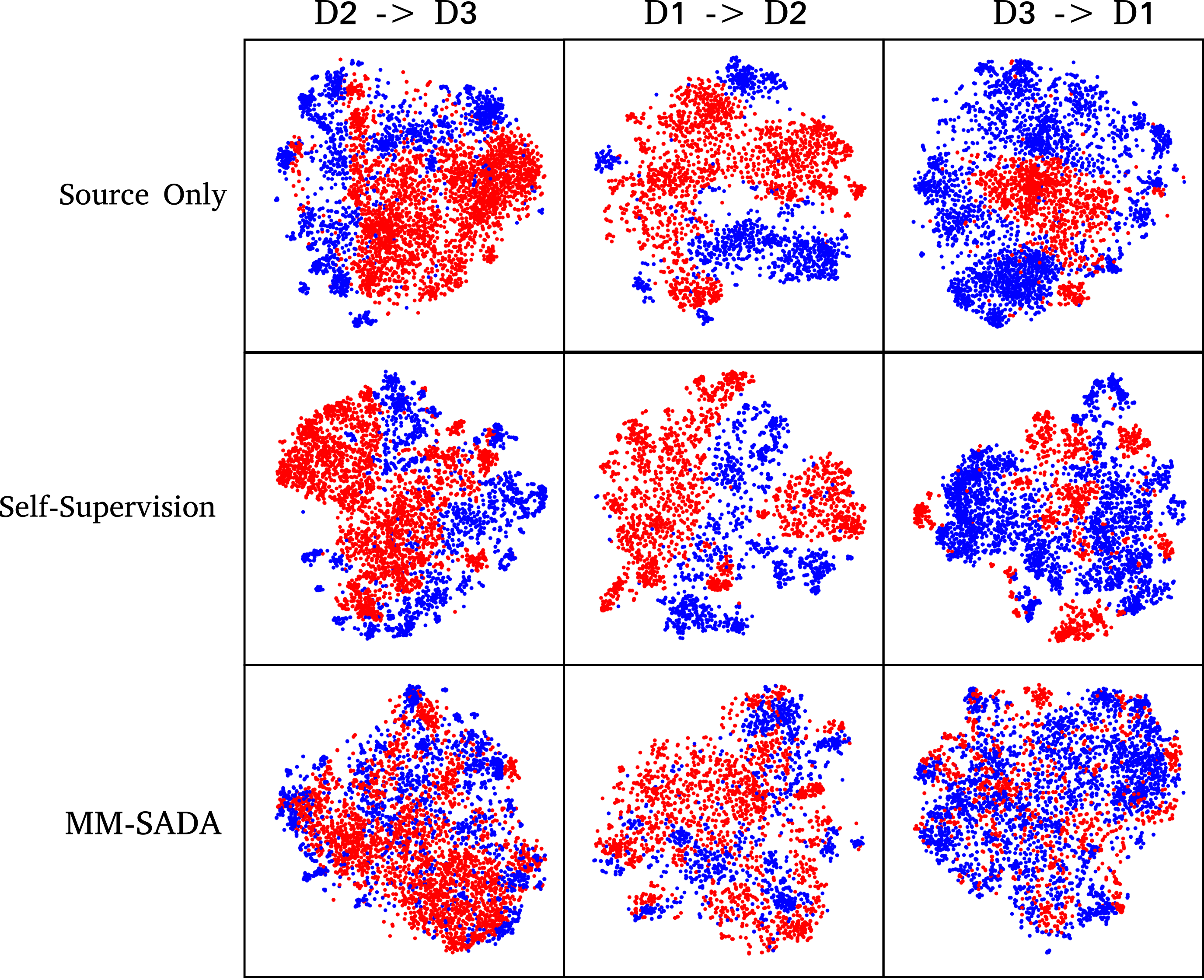}
      %  \caption{RGB}
    \end{subfigure}
    \hspace*{0.02\textwidth}
    \begin{subfigure}[]{0.48\textwidth}
        \includegraphics[width=\textwidth]{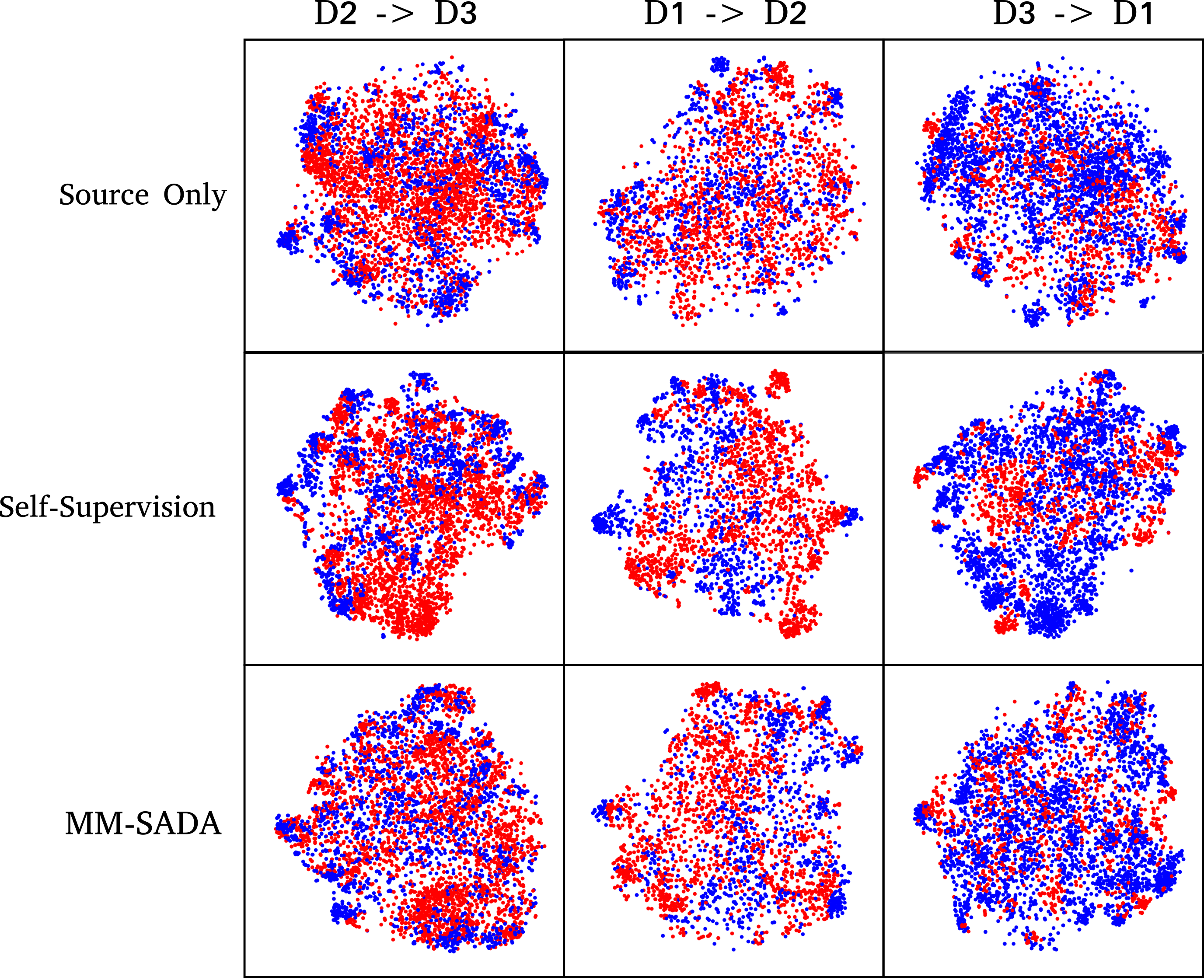}
      %  \caption{Flow}
    \end{subfigure}
 %   \vspace*{-8pt}
    \caption{t-SNE plots of RGB (left) and Flow (right) feature spaces produced by source-only, self-supervised alignment and our proposed model MM-SADA. \target{Target} is shown in red and {\color{blue}source} in blue. Our method better aligns both modalities.}
    \vspace*{-6pt}
    \label{fig:feature_plots}
\end{figure*}

\subsection{Ablation Study}
\label{sec:ablation}
\vspace*{-6pt}

Next, we compare the individual contributions of different components of MM-SADA. We report these results in Table~\ref{tab:Ablation}. The self-supervised component on its own gives a $2.4\%$ improvement over no adaption. This shows that self-supervision can learn features common to both source and target domains, adapting the domains. Importantly, this on average outperforms the three baselines in Table~\ref{tab:Compare}. Adversarial alignment per modality gives a further $2.4\%$ improvement as this encourages the source and target distributions to overlap, removing domain specific features from each modality. Compared to adversarial alignment only, our method improves in 5 of the 6 domains and by up to 3.2\%.

For the single pair noted earlier, $D3 \rightarrow D2$, self-supervision alone outperforms source-only and all other methods reported in Table~\ref{tab:Compare} by 1.1\%. 
%This shows the self-supervision method's potential to solve even the hardest cases, 
However when combined with domain adaptation using $\lambda_d = 1$, the overall performance of MM-SADA reported in Table~\ref{tab:Compare} cannot beat the baseline. In Table~\ref{tab:Ablation}, we show that when halving the contribution of adversarial component to $\lambda_d=0.5$,
%changing the contribution of adversarial to $0.5\times \lambda_d$,%
MM-SADA can achieve 56.9\% outperforming the source-only baseline. Therefore self-supervision can improve performance where marginal alignment domain adaptation techniques fail. %Therefore self-supervision can be effective as a domain adaptation technique in cases of large domain shift. 

Figure~\ref{fig:robustness} plots the performance of MM-SADA as $\lambda_d$ changes. Note that $\lambda_c$ can be chosen by observing the performance of self-supervision on source-domain labels, while $\lambda_d$ requires access to target data. We show that our approach is robust to various values of $\lambda_d$, with even higher accuracy at $\lambda_d=0.75$ than those reported in Table~\ref{tab:Ablation}.

Table ~\ref{tab:ModalityAblation} shows the impact of our method on the performance of the modalities individually. Predictions are taken from each modality separately before late fusion. RGB, the less robust modality, benefits most from MM-SADA, improving over source-only by $6.5\%$ on average, whereas Flow improves by $1.6\%$. The inclusion of multi-modal self-supervision provides $2.4\%$ and $1.1\%$  improvements for RGB and Flow, compared to only using adversarial alignment. This shows the benefit of employing self-supervision from multiple modalities during alignment. 
%\jonny{We compare the impact of MM-SADA on the performance of both RGB and Flow streams in Table \ref{tab:Modalities}, we evaluate the flow streams of our source-only model and MM-SADA using the corresponding individual classifier $G^m$. RGB stream benefits most from our alignment techniques, with a $6.5\%$ improvement over the source model. Flow does not increase as much from the source-only model to MM-SADA due to the greater robustness to environmental changes, however there is still a $1.6\%$ improvement. This shows the benfits of aligning both modalities.}

We also compare two approaches for multi-modal self-supervision in Table~\ref{tab:AblatSynch}. The first, which has been used to report all results above, learns the correspondence of RGB and Flow within the same action segment. We refer to this as \textit{`Seg. Corr.'}. The second learns the correspondence only from time-synchronised RGB and Flow data, which we call \textit{`Sync'}.
%To evaluate the self-supervision we compare to \textbf{Complete-Synchronous alignment}, where positive synchronised examples are sampled from the same frame. This is in contrast to our proposed method where positive examples are sampled from the same segment but not necessarily from the corresponding frames.
The two approaches are comparable in performance overall, with no difference on average over the domain pairs. This shows the potential to use a number of multi-modal self-supervision tasks for alignment.
%Our approach is to sample examples from each modality at different temporal locations within the same segment, the alternative is to sample at the same temporal location. 
%Notice that sampling at the same temporal location, complete synchronisation, performs worse than our proposed method. This is due to the egocentric nature of the EPIC Kitchens dataset, the network is able to learn camera motion to solve the self-supervised task. In comparison with our method, that sample modalities from different frames in a segment, the network cannot simply use camera motion to predict correspondence of modalities. Our sampling method requires understanding of how the visual environment relates to motion that compliments action recognition, with the inability to focus directly on motion artifacts to solve the task. 

\vspace*{-4pt}
\subsection{Qualitative Results}
\label{sec:qualitative}
\vspace*{-4pt}

% Figure~\ref{fig:qualitative} shows qualitative results of our method relative to source-only performance, with three success cases and one failure case for two pairs of domains. Without adaptation, models cannot utilise appropriate visual cues in the target environment, \ie{} appearance of chopping board and knife or sink and tap, therefore the model fails to predict cut and wash. Both adapted and non-adapted models struggle with ambiguous examples where different actions are occurring using both hands.

Figure~\ref{fig:feature_plots} shows the t-SNE~\cite{vanDerMaaten2008} visualisation of the RGB~(left) and Flow (right) feature spaces $F^m$. Several observations are worth noting from this figure. First, Flow shows higher overlap between source and target features pre-alignment (first row). This shows that Flow is more robust to environmental changes. Second, self-supervision alone (second row) changes the feature space by separating the features into clusters, that are potentially class-relevant. This is most evident for $D3 \rightarrow D1$ on the RGB modality (second row third column). However, alone this feature space still shows domain gaps, particularly for RGB features. Third, our proposed MM-SADA (third row) aligns the marginal distributions of source and target domains. % It is clear that Flow is more robust to environmental changes due to larger overlap in source and target domains than RGB. Self-supervision helps to separate the target domain into class clusters, this is particularly visible for $D3 \rightarrow D1$ on the RGB modality. Finally we show that are method aligns the marginal distributions of source and target domains.

\section{Conclusion and Future Work}

We proposed a multi-modal domain adaptation approach for fine-grained action recognition utilising multi-modal self-supervision and adversarial training per modality. We show that the self-supervision task of predicting the correspondence of multiple modalities is an effective domain adaptation method. On its own, this can outperform domain alignment methods~\cite{long2015learning,saito2018maximum}, by jointly optimising for the self-supervised task over both domains. Together with adversarial training, the proposed approach outperforms non-adapated models by $4.8\%$. We conclude that aligning individual modalities whilst learning a self-supervision task on source and target domains can improve the ability of action recognition models to transfer to unlabelled environments. 

Future work will focus on utilising more modalities, such as audio, to aid domain adaptation as well as exploring additional self-supervised tasks for adaptation, trained individually as well as for multi-task self-supervision.

\noindent \textbf{Acknowledgement} Research supported by EPSRC LOCATE (EP/N033779/1) and EPSRC Doctoral Training Partnershipts (DTP). The authors acknowledge and value the use of the ESPRC funded Tier 2 facility, JADE.
%Self-supervision can utilise the multi-modal nature of video for action recognition, and adapt models to unlabelled target environments. By predicting the correspondence of multiple modalities on both the source and target domains the network can learn features useful not only for source classification by that is generalise to the target domain. By combining self-supervision with adversarial domain alignment, our method can actively remove domain specific features to align the marginal statistics of both and source and target domains. Adversarial alignment has been shown in this work to be more robust to the class imbalance in the EPIC Kitchens dataset than other discrepancy minimisation techniques, making it more suited to real world situations.\\

\clearpage
{\small
\bibliographystyle{ieee_fullname}
\bibliography{egbib}
}

\end{document}